\documentclass{article}
\usepackage{spconf,amsmath,graphicx,nccmath}
\usepackage{multicol}
\usepackage[figurename=Fig.]{caption}
\usepackage{subcaption}% Example definitions.
\title{Visualisation of Medical Image Fusion and Translation for Accurate Diagnosis of High Grade Gliomas}
\name{Nishant Kumar$^1$, Nico Hoffmann$^2$, %Martin Oelschl$\ddot{a}$gel$^3$, Edmund Koch$^3$,%
Matthias Kirsch$^3$,
Stefan Gumhold$^1$\thanks{This work was supported by the European Social Fund (project no. 100312752) and the Saxonian Ministry of Science and Art.}}
\address{$^1$Computer Graphics and Visualisation, Technische Universit$\ddot{a}$t Dresden, Germany \\ $^2$Helmholtz-Zentrum Dresden-Rossendorf, Germany %\\ $^3$Clinical Sensoring and Monitoring, Technische Universit$\ddot{a}$t Dresden, Germany 
\\ $^3$Department of Neurosurgery, University Hospital Carl Gustav Carus, Germany}

\begin{document}
%\ninept
%
\maketitle
\begin{abstract}
The medical image fusion combines two or more modalities into a single view while medical image translation synthesizes new images and assists in data augmentation. Together, these methods help in faster diagnosis of high grade malignant gliomas. However, they might be untrustworthy due to which neurosurgeons demand a robust visualisation tool to verify the reliability of the fusion and translation results before they make pre-operative surgical decisions. In this paper, we propose a novel approach to compute a confidence heat map between the source-target image pair by estimating the information transfer from the source to the target image using the joint probability distribution of the two images. We evaluate several fusion and translation methods using our visualisation procedure and showcase its robustness in enabling neurosurgeons to make finer clinical decisions. 
\end{abstract}
\section{Introduction}
\label{sec:intro}
High grade malignant gliomas such as anaplastic astrocytoma and glioblastoma multiforme (GBM) are some of the most aggressive brain tumors having rapid growth tendencies. Thus, a non-invasive pre-operative clinical examination of the human subject is done by medical professionals using various imaging techniques to carefully estimate the location and size of the tumor. The outcome of this procedure is especially important since neurosurgeons wants to preserve as much healthy tissues as possible during surgical interventions. 

\begin{figure}[htb]
\centering
\centerline{\includegraphics[width=7 cm]{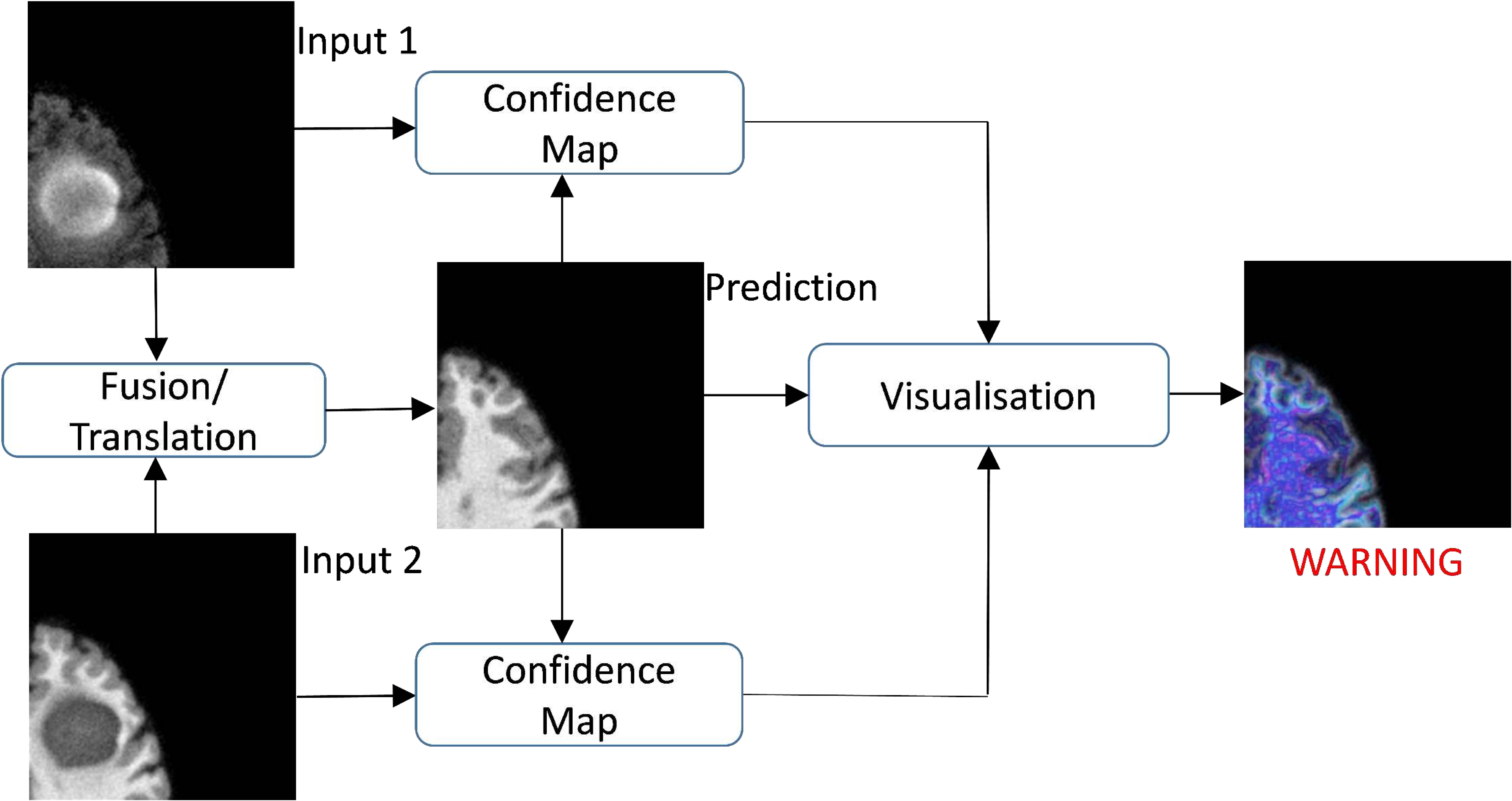}}
\caption{Our proposed visualisation framework: there are two input image pairs fed to fusion or translation algorithms, a predicted image, confidence maps and visualisation results.}\medskip
\end{figure}

Imaging modalities such as MRI provide high resolution anatomical information of the brain. However, it relies solely on morphological criteria to characterize malignant tissues, revealing no functional information like glucose metabolism provided by modalities such as PET. The anatomical information together with functional information is crucial to establish a surgical decision about tumor resection. 
The post-hoc medical image fusion of MRI and PET image pairs combines anatomical and functional information and therefore provides faster diagnosis. However, the neurosurgeons have low credence on such fusion methods since either these methods are highly intricate $\cite{A3,A4,A5,A6}$ or are blackboxes like deep learning based methods $\cite{P1,A2}$ with low explainability. Secondly, since there is no gold standard for an ideal fused image, all of these fusion methods evaluate the quality of the results based on some metric scores $\cite{P4,P5,A7}$. This kind of evaluation is not useful since surgeons require visual insights into the quality of fused image. Overall, these fusion methods are impractical for real time use in surgical planning and interventions.
Moreover, there is an additional challenge of missing data since the pre-operative MRI acquisition is either done with T1 or T2 relaxation times due to which the underlying anatomical features are not revealed completely. Recently, Generative adversarial network (GAN) based methods such as CycleGAN $\cite{P19}$ and Conditional GAN $\cite{P21}$ have been widely popular to synthesize translated medical images from a given source domain e.g. MR-T2 to a target domain e.g. MR-T1. However, $\cite{P17}$ showed that these methods introduce hallucinated features in the target image if the network is trained with over or under representation of target domain class (e.g. w or w/o tumor). Due to this, it is not recommended for neurosurgeons to rely on these translation methods for medical diagnosis $\cite{P17}$. Also, there are concerns of a legal challenge of an objectionable machine decision in sensitive cases such as gliomas especially when there is no tool available that helps to visualise the trustability of these fusion and translation algorithms. 

Interestingly, there have been techniques proposed which attempts to visualise the prediction of blackbox neural networks. Gradient based explainable algorithms $\cite{P6,P7,P8}$ and relevance score based methods $\cite{P9, P10}$ provides good visual explanation of the model outputs but requires either the backpropagation heuristics along the layers of a neural network or gradient computation of the intermediate layers and activation functions. Hence, they are only applicable to neural network specific methods. Perturbation based visualisation $\cite{P11,P12, P13, P14}$ edits the the pixel intensity of the input image with some noise like blurring or occlusion and the change in the prediction probability of the output is observed. Therefore, this method could be applied to any blackbox fusion/translation algorithm. However, it needs several feed forwards thereby making them slow, expensive and unfit for real time deployment. Secondly, the applicability of such methods on unusual artifacts such as speckle noise which are quite common in medical images remains unexplored.

Lastly, all the above visualisation methods have been developed keeping classification problem in mind where the task is to detect an object in an image not necessarily from medical domain. However, in a visualisation approach for a black box medical image fusion or translation algorithm, the aim is to compute the confidence of each pixel of the predicted target image based on the amount of information transfer from a given source image. Therefore, the main contribution of this work is to develop a novel visualisation technique to compute a confidence heat map on a source-target image pair in order to recognize trustable regions in the target image. Our method could be applied for learning as well as non-learning based fusion and translation methods and has real time applications in surgical planning.

\section{METHOD}
\label{sec:format}
We take the grayscale source and target image patches of size $W$ and convert them into one dimensional feature vectors. Assuming source feature vector as a discrete and independent random variable $X$ with marginal probability distribution function (MPDF) $f_X(x)$ and target feature vector as a discrete and dependent random variable $Y$ with MPDF $f_Y(y)$, the goal is to model a joint probability distribution function (JPDF) $f_{X,Y}(x,y)$. However, the estimation of JPDF $f_{X,Y}(x,y)$ given the individual MPDFs $f_X(x)$ and $f_Y(y)$ is an ill-posed inverse problem with many possible solutions. Although the joint cumulative distribution function (JCDF) $F_{XY}(x,y)= P(X \leq x, Y \leq y)$ of random variables $X$ and $Y$ is unknown, the individual marginal cumulative distribution functions (MCDF) of the random variables are given by $F_{X}(x) = P(X \leq x) = \sum_{x}f_X(x)$ and $F_{Y}(y) = P(Y \leq y) = \sum_{y}f_Y(y)$. Also, there are minimum and maximum correlations between $X$ and $Y$ that satisfies $F_{XY}^{L}(x,y) \leq F_{XY}(x,y) \leq F_{XY}^{U}(x,y)$ where $F_{XY}^{L}(x,y)$ and $F_{XY}^{U}(x,y)$ are upper and lower boundaries of $F_{XY}(x,y)$ which could be computed using Fr$\acute{e}$chet inequalities criteria. Now, given the respective MCDFs and boundary JCDFs of the two random variables, we compute the boundary covariances using Hoeffding$'$s covariance identity as:\useshortskip
\begin{equation}
\sigma^{L,U}_{x,y} = \sum_{x}\sum_{y} (F_{XY}^{L,U}(x,y) - F_{X}(x)F_{Y}(y))
\end{equation}
Based on the above equation, we define the Pearson correlation coefficients of upper and lower boundaries as $\rho^{L} = \sigma^{L}_{x,y}/\sigma_{x}\sigma_{y}$ and $\rho^{U} = \sigma^{U}_{x,y}/\sigma_{x}\sigma_{y}$. Fr$\acute{e}$chet inequalities also holds for covariances and correlation coefficients meaning $\sigma^{L}_{x,y} \leq \sigma_{x,y} \leq \sigma^{U}_{x,y}$ and $\rho^{L} \leq \rho \leq \rho^{U}$. Assuming $f_{X,Y}^{L}(x,y)$ and $f_{X,Y}^{U}(x,y)$ as the lower and upper bounds JPDFs, then according to $\cite{P15}$, we can model the  $f_{X,Y}(x,y)$ of our concerned discrete bivariate distribution as:\useshortskip

\begin{equation}
\resizebox{\hsize}{!}
{
$
f_{X,Y}(x,y) =
\begin{cases}
    \frac{\rho}{\rho^{U}} f_{X,Y}^{U}(x,y) + (1-\frac{\rho}{\rho^{U}}) f_X(x) f_Y(y) , & \text{if } \rho > 0\\
    \frac{\rho}{\rho^{L}} f_{X,Y}^{L}(x,y) + (1-\frac{\rho}{\rho^{L}}) f_X(x) f_Y(y), & \text{if } \rho \leq 0
\end{cases}
$
 
}
\end{equation}

With $f_{X,Y}(x,y)$, $f_{X}(x)$ and $f_{Y}(y)$ of source and target features known, the amount of information which target feature vector contains about source feature vector could be calculated using the method proposed in $\cite{P3}$.
%\begin{equation}
%M_{X,Y}(x,y) =  %\sum_{x,y}f_{X,Y}(x,y) %log_2\frac{f_{X,Y}(x,y)}{f_{X}(x)f_5{Y}(y)}   
%\end{equation}
However, this approach computes pixel wise information with $W = 1$ between source and target image thereby excluding the neighborhood information. Additionally, the final mutual information scores between the target image and each source image are aggregated which means that the two source images are not measured at the same scale. This results in biased decisions towards the source image with the highest entropy and consequently non-trustworthy fusion and translation quality assessment. Given the sensitivity of assessing high grade gliomas, we select a higher patch size of $W=7$ and  include individual entropies in the mutual information calculation to negate the scalability issue of the source images. Our patch level normalised confidence score $S_{X,Y}(x,y)$ is given by:\useshortskip
\begin{equation}
S_{X,Y}(x,y) = \frac{2\sum_{x,y}f_{X,Y}(x,y) log_2\frac{f_{X,Y}(x,y)}{f_{X}(x)f_{Y}(y)}}{\sum_{x}f_{X}(x)log_2f_{X}(x) + \sum_{y}f_{Y}(y)log_2f_{Y}(y)}
\end{equation}

where $S_{X,Y}(x,y) \epsilon [0,1]$ with values near to 1 conveys pixels with high confidence of information transfer from source image to the predicted target image.

\begin{figure*}[!htb]
\centering
\includegraphics[width=\textwidth]{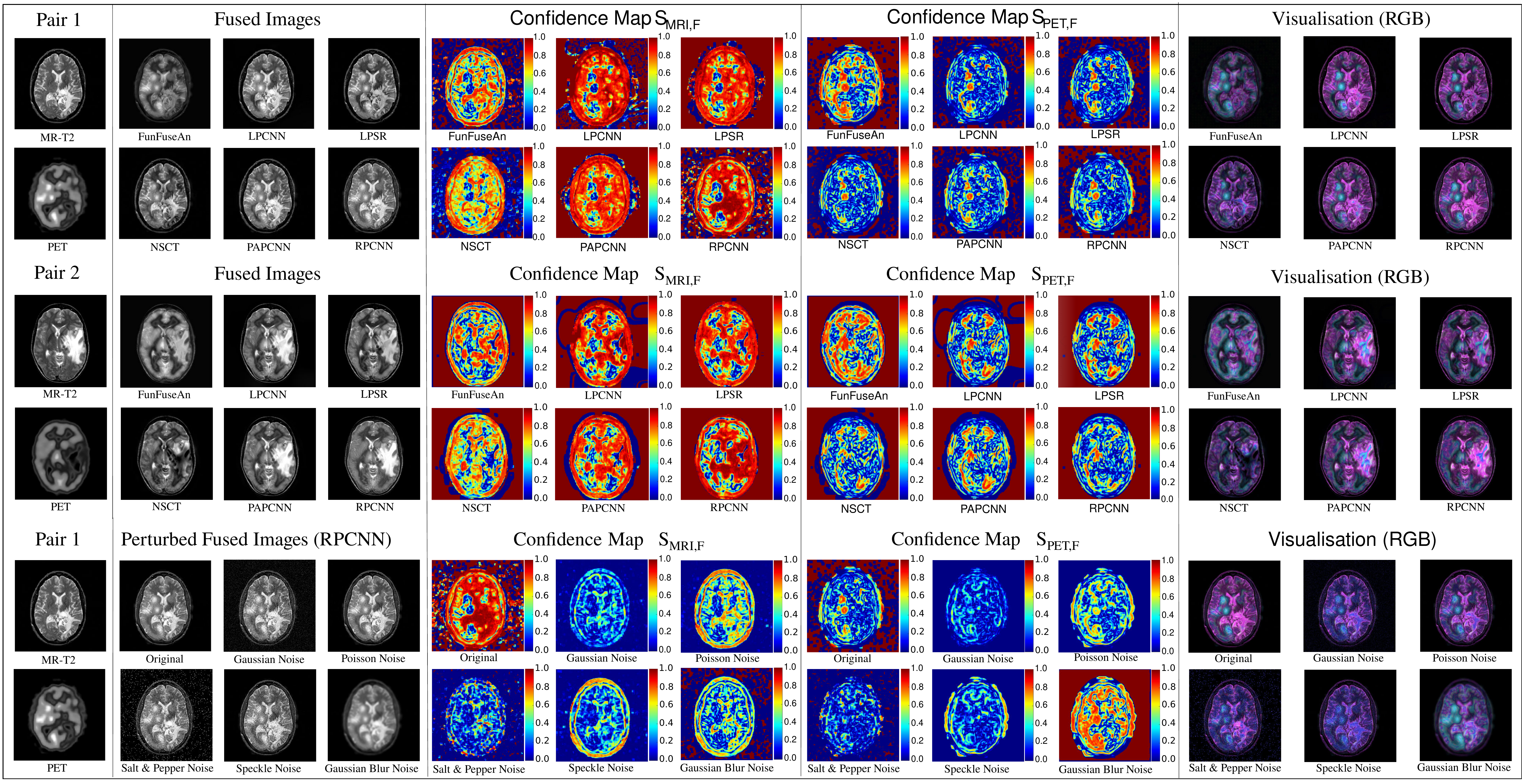}
\caption{Fusion results of our visualisation framework: modalities MR-T1 and PET are the inputs to six different fusion methods which generates respective fused images. Then, the $S_{MRI,F}$ and $S_{PET,F}$ confidence maps are computed between fused images and each of the inputs. Eventually, the fused images are evaluated for its reliability using our fusion visualisation settings.}
\end{figure*}

\section{Experimental results and analysis}
\label{sec:pagestyle}
\subsection{Fusion and translation visualisation settings}
For fusion settings, we acquired several pre-registered publicly available MR-T2 and PET-FDG image pairs of unique human subjects from
Harvard  Whole  Brain  Atlas $\cite{P18}$ with subjects suffering from different forms of high grade glioma such as grade III astrocytoma and grade IV GBM. All the subjects were aged between 35-75 years among both genders and all the image pairs were analyzed as axial slices
with a voxel size of 1.0 x 1.0 x 1.0
mm and tumor tissues clearly visible. We applied our visualisation approach for the evaluation of six different state-of-the-art post-hoc MRI-PET fusion algorithms from recent past. Two of them were convolutional neural network based methods namely LPCNN $\cite{P1}$ and FunFuseAn $\cite{A2}$ whereas others were non-learning based methods including nonsubsampled
contourlet transforms NSCT $\cite{A3}$ and RPCNN $\cite{A6}$, combination
of multi-scale transform and sparse representation LPSR $\cite{A4}$ and  nonsubsampled
shearlet transform PAPCNN $\cite{A5}$. We defined $S_{MRI,F}$ as the confidence heat map between the fused image $I_F$ and the source MRI image, $S_{PET,F}$ as the confidence heat map between $I_F$ and the source PET image. We color $I_F$ by defining RGB channels as $R = \alpha S_{MRI,F} + (1-\alpha) I_F$, $G = \alpha S_{PET,F} + (1-\alpha) I_F$ and $B = I_F$ where $\alpha = 0.7$ is the color intensity parameter. Now, according to our defined RGB model, we expect magenta  color $(1,0,1)$ in regions of the fused image with $S_{MRI,F} \approx 1$ while cyan color $(0,1,1)$ in regions with $S_{PET,F} \approx 1$. In addition to the above evaluation, we perturb the fused image of RPCNN method by white gaussian noise $N_G \sim \mathcal{N} (0, 0.01)$, poisson noise $N_P \sim P(x)$, salt and pepper noise with noise density of 0.05, speckle noise $I_F = I_F + \mathcal{N} (0, 0.05)*I_F$ and blur noise with a 2-D Gaussian smoothing kernel with standard deviation of 0.5 to evaluate the change in the confidence heat maps.

For translation settings, we used the publicly available BRATS 2013 dataset containing MR-T2 Flair (source domain) and MR-T1 (target domain) images and then visualised the confidence of CycleGAN $\cite{P19}$, CondGAN $\cite{P21}$ and L1 based translation methods by following the training and testing settings given in $\cite{P17}$ for 3 different percentages of training data containing tumor ranging from $0-100\%$. Assuming the source MR-T2 Flair image as $I_{T2}$ and the target MR-T1 image as $I_{T1}$, then we color the predicted target MR-T1 image $I_{\widehat{T1}}$ by defining RGB channels as $R = \alpha S_{T2} + (1-\alpha) I_{\widehat{T1}}$, $G = \alpha S_{T1} + (1-\alpha) I_{\widehat{T1}}$ and $B = I_{\widehat{T1}}$ where $S_{T2}$ is the confidence heat map between $I_{\widehat{T1}}$ and $I_{T2}$,  $S_{T1}$ is the confidence heat map between $I_{\widehat{T1}}$ and $I_{T1}$ and $\alpha = 0.7$. Since a robust translation method should result in  $I_{\widehat{T1}} \approx I_{T1}$ and $I_{\widehat{T1}} \neq I_{T2}$, there should be very low confidence between $I_{\widehat{T1}}$ and $I_{T2}$ with $S_{T2} \approx 0$ and pretty high confidence between $I_{\widehat{T1}}$ and $I_{T1}$ with $S_{T1} \approx 1$. Therefore, cyan (0,1,1), blue (0,0,1) and magenta (1,0,1) reveals best to worst performances in that order.

\subsection{Visual results of fusion and translation algorithms}
The first and second set of Fig. 2. shows the confidence heat maps $S_{MRI,F}$, $S_{PET,F}$ and visualisation results of various fusion methods on two MRI-PET image pairs. $S_{MRI,F}$ of the fusion methods convey that RPCNN has highest confidence in preserving MRI features but has lower confidence in preserving PET features as well as background regions due to unwanted noise. The methods like LPSR and PAPCNN also performs well in preserving MRI features and has higher confidence for background regions. The analysis of $S_{PET,F}$ reveals that FunFuseAn performs best compared to all other methods to preserve PET features in the fused image. The visualisation (RGB) results convey that RPCNN has strong magenta color for the MRI features while FunFuseAn has strong cyan color representation around the regions with PET features. This validates the results in the confidence heat maps where RPCNN and FunFuseAn performed better than other fusion algorithms in preserving MRI and PET features respectively. The third set of results in Fig. 2. shows the decrease in confidence of the fused image for almost all the heat maps after it was perturbed with various types of noises, conveying that the addition of noise leads to loss of information transfer from source images. Interestingly, adding some gaussian blur noise leads to increase in confidence of the heat map $S_{PET,F}$ which could be explained by the fact that input PET image is of lower resolution and blurry compared to input MRI image.  

\begin{figure}[!htb]
\centering
\centerline{\includegraphics[width=8 cm,scale=0.5]{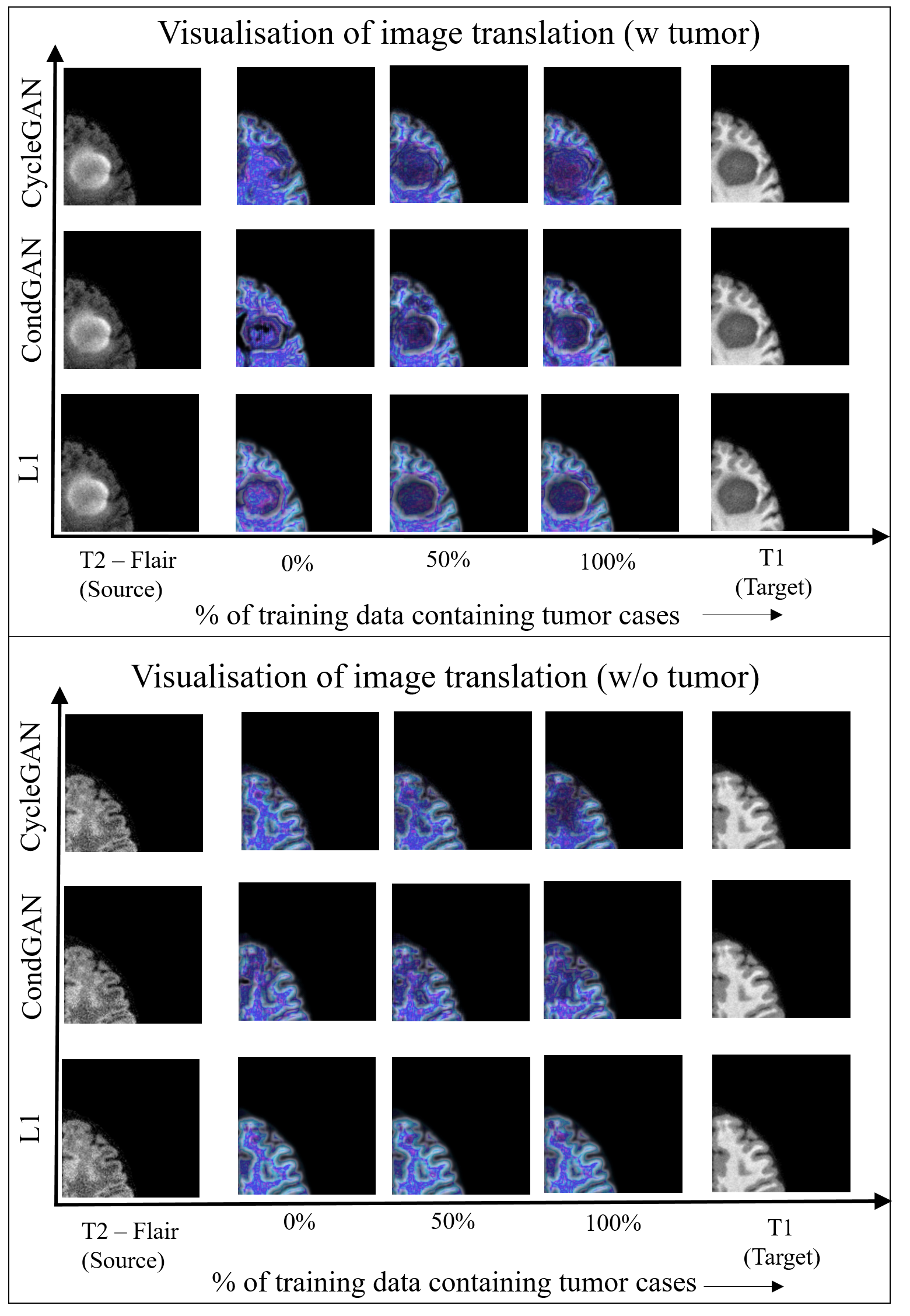}}
\caption{Translation results of our visualisation framework: The first and second set of images illustrate the MR-T2 Flair and MR-T1 image pair for tumor and non-tumor test cases along with the visualisation results of 3 different training data cases respectively.}\medskip
\end{figure}

The first set of Fig. 3. reveals that with $0\%$ tumor cases, CycleGAN has far greater and widespread blue regions compared to CondGAN and L1 based translation methods while there are some magenta regions in small patches and negligible cyan colored pixels. This means that CycleGAN results in very low confidence between $S_{\widehat{T1}}$ and $S_{T1}$ among the pixels colored with magenta or blue. L1 based translation method on the other hand has less blue or magenta regions and contains several regions colored with cyan. With the cases $50\%$ and $100\%$, the blue and magenta regions in the CycleGAN and CondGAN methods decreases, however, L1 method has greater cyan regions compared to blue and magenta regions. Hence, for the tumor test case, the predicted target image $I_{\widehat{T1}}$ from L1 loss is more reliable than GAN based methods irrespective of the percentage of training data containing tumor. CycleGAN and to some extent CondGAN performs poorly in information transfer leading to wiping out of tumor features especially when there is under-representation of tumor cases in the training set. The second set of Fig. 3. shows that with $0\%$ tumor cases, L1 based translation method again performs better than the other methods although CycleGAN comes in second position with less blue or magenta regions compared to CondGAN. However, as the percentage of tumor cases in training data is increased, CycleGAN performs worse with greater blue and magenta colored regions compared to CondGAN. L1 method is not affected by the change in case and maintains the amount of pixels colored with cyan, blue and magenta. This conveys that L1 loss is again more reliable compared to CycleGAN and CondGAN with non-tumor test case, as the latter methods hallucinate the predicted target images by adding tumor features into them when there is over-representation of tumor cases in the training data.

%\begin{table}
%\small
%\centering
%\caption{Assessment of fusion methods based on our visualisation approach}\label{tab1}
%\begin{tabular}{|p{0.69cm}|p{1.31cm}|p{1.0cm}|p{1.0cm}|p{0.8cm}|p{0.88cm}|}
%\hline Score & Method & Sens.(\%) & Spec.(\%) & F1(\%) & Acc.(\%) \\
%\hline
%          &  LPCNN & 99.49  & 93.70 & 96.69 & 96.14 \\
%          &  RPCNN & 93.54  & 99.70 & 96.32 & 97.11 \\
%$S_{MRI}$ &  FunFuseAn & 95.44  & 81.83 & 90.43 & 87.55 \\
%          &  PAPCNN & 97.89  & 99.08 & 98.29 & 98.58 \\
%          &  LPSR & 98.67  & 99.87 & 99.22 & 99.36 \\
%          &  NSCT & 98.37  & 98.95 & 98.46 & 98.70 \\
%\hline
%          &  LPCNN & 94.72  & 82.46 & 88.99 & 87.62 \\
%          &  RPCNN & 98.04  & 99.63 & 98.75 & 98.96 \\
%$S_{PET}$ &  FunFuseAn & 98.60  & 81.23 & 91.16 & 87.80 \\
%          &  PAPCNN & 95.92  & 98.34 & 96.76 & 97.33 \\
%%          &  LPSR & 95.18  & 99.52 & 97.12 & 97.69 \\
%          &  NSCT & 94.64  & 98.13 & 95.90 & 96.66 \\
%\hline
%          &  LPCNN & 97.69  & 88.69 & 93.41 & 92.48 \\
%          &  RPCNN & 94.95  & 99.64 & 96.99 & 97.66 \\
%$S_{AV}$  &  FunFuseAn & 94.98  & 80.82 & 85.36 & 80.42 \\
%          &  PAPCNN & 96.24  & 98.43 & 96.90 & 97.50 \\
%          &  LPSR & 96.92  & 99.70 & 98.11 & 98.53 \\
%          &  NSCT & 95.84  & 98.18 & 96.54 & 97.20 \\
%\hline
%\end{tabular}
%\end{table}//

\section{Conclusion and DISCUSSION}
\label{sec:typestyle}
In this work, we proposed a first of its kind visualisation tool to interpret the quality of medical image fusion and translation algorithms. One important application of our tool is that clinicians could visualise the confidence scores of the malignant regions of the brain such as high grade gliomas. We have presented key visual evidences that some of the evaluated algorithms performs better in preserving information in these specific regions compared to the other methods. Therefore, these methods should be cautiously used for interpretation by the clinicians in order to prevent any erroneous diagnostic decisions. In future, we plan to apply a kernel density estimate on the input and the target feature vectors and evaluate the response of our visualisation approach.

\bibliographystyle{IEEEbib}
\bibliography{strings,refs}

\end{document}